\documentclass[runningheads]{llncs}
\usepackage{graphicx}
\usepackage{amsmath,amssymb} 
\usepackage{color}
\usepackage{float}
\usepackage{array}
\newcolumntype{C}[1]{>{\centering\let\newline\\\arraybackslash\hspace{0pt}}m{#1}}

\usepackage[width=122mm,left=12mm,paperwidth=146mm,height=193mm,top=12mm,paperheight=217mm]{geometry}
\begin{document}
\pagestyle{headings}
\mainmatter

\title{Attributes for Improved Attributes: A Multi-Task Network for Attribute Classification} 
\titlerunning{Attributes for Improved Attributes}


\author{Emily M. Hand and Rama Chellappa}
\institute{University of Maryland, College Park}

\maketitle

\begin{abstract}

Attributes, or semantic features, have gained popularity in the past few years in domains ranging from activity recognition in video to face verification. Improving the accuracy of attribute classifiers is an important first step in any application which uses these attributes. In most works to date, attributes have been considered to be independent. However, we know this not to be the case. Many attributes are very strongly related, such as \emph{heavy makeup} and \emph{wearing lipstick}. We propose to take advantage of attribute relationships in three ways: by using a multi-task deep convolutional neural network (MCNN) sharing the lowest layers amongst all attributes, sharing the higher layers for related attributes, and by building an auxiliary network on top of the MCNN which utilizes the scores from all attributes to improve the final classification of each attribute. We demonstrate the effectiveness of our method by producing results on two challenging publicly available datasets.


\keywords{Attributes, Face Recognition, CNN, Multi-Task Learning}
\end{abstract}

\section{Introduction}

Attributes are mid-level representations used for the recognition of activities, objects, and people \cite{Duan12} \cite{Zheng14} \cite{Zhang14}. Attributes provide an abstraction between the low-level features and the high-level labels. They have seen the most success in face recognition and verification \cite{Kumar09} \cite{Kumar11}. In the face recognition domain, attributes include \emph{gender}, \emph{race}, \emph{age}, \emph{hair color}, \emph{facial hair}, etc. These semantic features are very intuitive, and they allow for much more understandable descriptions of objects, people, and activities. Reliable estimation of facial attributes is useful for many different tasks. HCI applications may require information about gender in order to properly greet a user (i.e. Mr. or Ms.) and other attributes such as expression in order to determine the mood of the user. Facial attributes can be used for identity verification in low quality imagery, where other verification methods may fail. Suspects are often described in terms of attributes, and so they can be used to automatically search for suspects in surveillance video. Attributes can be used to search a database of images very quickly. They have been used very successfully in image search and retrieval in the past few years \cite{Kumar09} \cite{Kumar11} \cite{Siddiquie11}. 

Improving the accuracy of attribute classifiers is a challenging problem in itself and has been of recent interest due to the release of several large-scale attribute datasets \cite{Liu15}. Convolutional neural networks (CNNs) have replaced most traditional methods for feature extraction in many computer vision problems \cite{Kriz12} \cite{Vinyals15}. They have proved to be effective in attribute classification as well \cite{Zhang14} \cite{Abdu15} \cite{Levi15}. However, with few exceptions, attributes have been treated as independent from each other. From a simple example - a woman wearing lipstick and earrings - we can see that this is not the case. If the subjects are wearing lipstick and earrings, the probability that they are women is much higher than if they did not exhibit those attributes, and the reverse is also true. Treating each attribute as independent fails to use the valuable information provided by the other attributes. Attributes fit nicely into a multi-task learning framework, where multiple problems are solved jointly using shared information \cite{Arg07} \cite{Para10} \cite{Caruana97}.

We propose a multi-task deep CNN (MCNN) with an auxiliary network (MCNN-AUX) on top in order to utilize information provided by all attributes in three ways: by sharing the lower layers of the MCNN for all attributes, by sharing the higher layers for similar attributes, and by utilizing all attribute scores from the MCNN in an auxiliary network in order to improve the recognition of individual attributes. We are able to achieve state-of-the-art performance on most attributes from two large-scale publicly available datasets: CelebA and LFWA \cite{Liu15}. 

The contributions of our work are as follows:
\begin{itemize}
	\item We develop a multi-task deep CNN for attribute classification.
	\item We develop an auxiliary network for MCNN which allows for explicit use of attribute scores to improve classification of other attributes.
	\item We demonstrate the effectiveness of our approach by evaluating on two challenging publicly available datasets - LFWA and CelebA.
	\item We achieve state-of-the-art performance for many attributes, some showing up to a $15\%$ improvement over other methods.
	\item We significantly decrease the number of parameters - over 4 times - and the amount of training time - over 16 times - required for the attribute classifier.
	\item Our method requires no expensive pre-training, alignment, or part extraction steps.
\end{itemize}

The remainder of the paper is organized as follows: Section \ref{sec:relatedwork} describes the related work on CNNs, multi-task learning, and attributes. Section \ref{sec:MCNN} discusses the MCNN architecture, and section \ref{sec:MCNN-AUX} describes the auxiliary network. In section \ref{sec:experiments} we outline the experiments we performed in order to test our methods as well as our results. Finally in section \ref{sec:conclusion} we discuss the impact of our work.

\section{Related Work} \label{sec:relatedwork}
There are large bodies of work on CNNs, Multi-Task Learning, and Attributes. We draw from all three areas to design the proposed method, MCNN-AUX. The relevant literature is reviewed in the following sections.

\subsection{CNN}
Deep CNNs have been widely used for feature extraction and have shown great improvement over hand-crafted features for many problems including object recognition, automatic caption generation, face detection, face recognition and verification, and activity recognition \cite{Girshick14} \cite{Kriz12} \cite{Vinyals15}. CNNs have quickly gained popularity since the introduction of open-source software tools which allow for straight-forward construction, training, and testing of deep CNNs. Caffe, Torch, and TensorFlow are among the most popular packages for implementing CNNs \cite{caffe}\cite{Abadi15}. The first big success for deep CNNs in a large-scale problem was in the 2012 Imagenet Challenge with a network that outperformed the then existing methods \cite{Kriz12}. Since then, a wide variety of CNN architectures have been proposed for many computer vision problems.

CNNs have also dominated the field of face recognition and verification. One of the most notable works in this domain is that of Deep-Face, which utilized a large dataset and applied both a siamese deep CNN and a classification CNN in order to maximize the distance between impostors and minimize the distance between true matches \cite{Taigman14}. Motivated by the success on the LFW dataset, researchers focused more on CNNs for face recognition and the networks became deeper and more complex \cite{Sun1} \cite{Sun2} \cite{Sun3} \cite{Sun4}.

In this work, we take advantage of the discriminative power of the CNN to learn semantic attribute classifiers as a mid-level representation for subsequent use in recognition and verification systems.

\subsection{Multi-Task Learning}
Multi-task learning (MTL) is a way of solving several problems at the same time utilizing shared information \cite{Arg07} \cite{Para10} \cite{Caruana97}. MTL has found success in the domains of facial landmark localization, pose estimation, action recognition, face detection, and many more \cite{Zhang14_2} \cite{Zhou13} \cite{Junho15} \cite{Zhang14_3} \cite{Devries14}.

In \cite{Wang09}, \cite{Wang10}, and \cite{Hwang11} attributes and object classes are learned jointly to improve overall object classification performance. \cite{Wang09} uses Multiple Instance Learning to detect and recognize objects in images by learning attribute-object pairs. \cite{Wang10} uses an undirected graph to model the correlation amongst attributes in order to improve object recognition. In \cite{Hwang11}, attributes and objects share a low-dimensional representation allowing for regularization of the object classifier. In our work, all attributes share the lower layers in the CNN, so that information common to all the attributes can be learned. Applying MTL to attribute prediction is very natural given the strong relationships among the facial attributes.

\subsection{Attributes}
Kumar et al. introduced the concept of attributes as image descriptors for face verification in \cite{Kumar09}. They used a collection of 65 binary attributes to describe each face image. They later extended this work with an addition of 8 attributes and applied their method to the problem of image search in addition to face verification \cite{Kumar11}.  Berg et al. created classfiers for each pair of people in a dataset and then used these classifiers to create features for a face verification classifier \cite{Berg12}. Here, rather than manually identifying attributes, each person was described by their likeness to other people. This is a way of automatically creating a set of attributes without having to exhaustively hand-label attributes on a large dataset. Prior to this, there were decades of research on gender and age recognition from face images \cite{Fu10}\cite{Ng12}.

CNNs have been used for attribute classification recently, demonstrating impressive results. Pose Aligned Networks for Deep Attributes (PANDA) achieved state-of-the-art performance by combining part-based models with deep learning to train pose-normalized CNNs for attribute classification \cite{Zhang14}. Focusing on age and gender, \cite{Levi15} applied deep CNNs to the Adience dataset. Liu et al. used two deep CNNs - one for face localization and the other for attribute recognition - and achieved impressive results on the new CelebA and LFWA datasets, outperforming PANDA on many attributes \cite{Liu15}. Unlike these methods, our MCNN-AUX requires no pre-training, alignment or part extraction.

Past work has generally considered attributes to be independent, with \cite{Kumar09}, \cite{Zhang14}, and \cite{Liu15} training a separate classifier for each attribute. \cite{Siddiquie11} uses the correlation amongst attributes to improve image ranking and retrieval. They use independently trained attribute classifiers and then learn pairwise correlations based on the outputs of these classifiers. Our method goes above and beyond this by training a single attribute network which classifies 40 attributes, sharing information throughout the network, and by learning the relationship among all 40 attributes, not just attribute pairs. \cite{Abdu15} used a multi-task network to learn attributes for animals and clothing, rather than faces. They utilize groupings as in \cite{Jaya14}, but they impose constraints at the feature level according to the groups. We incorporate groupings directly into the network by sharing layers amongst attributes in a single grouping.

\section{Multi-Task CNN}
\label{sec:MCNN}
The proposed MCNN takes an image as input and outputs 40 separate attribute scores, which are then thresholded to obtain binary outputs. We describe the details of the architecture below.

\subsection{ Architecture}
\begin{figure}[h]
\centering
\includegraphics[width=0.8\textwidth]{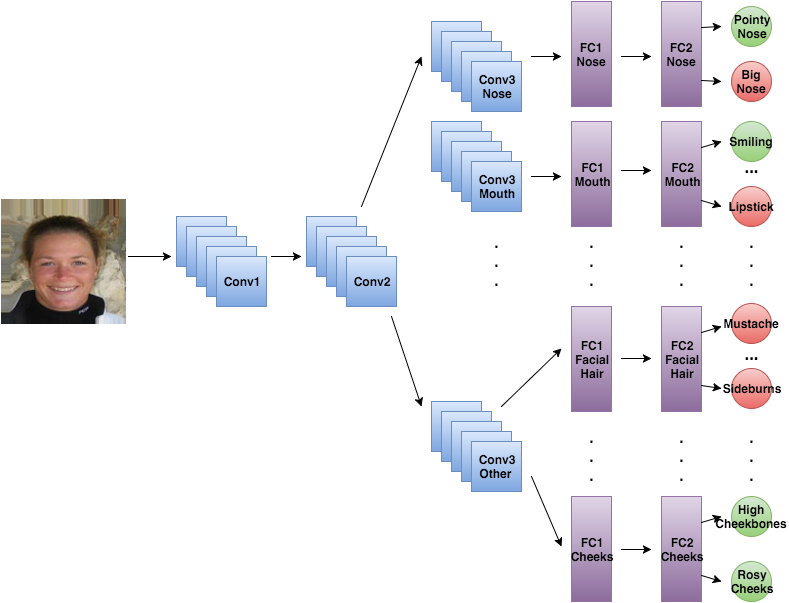}
\caption{Overview of MCNN. The input image on the left is cropped to 227x227 and the training mean is subtracted. The image is then passed through the convolution layers and the fully connected layers to produce attribute scores. The attribute scores are then thresholded to give a yes or no answer. The red attributes indicate a lack of the attribute and the green attributes indicate a positive instance.}
\label{fig:multi-taskcnn}
\end{figure}

Figure \ref{fig:multi-taskcnn} shows the MCNN architecture. Conv1 consists of 75 7x7 convolution filters, and it is followed by a ReLU, 3x3 Max Pooling, and 5x5 Normalization. Conv2 has 200 5x5 filters and it is also followed by a ReLU, 3x3 Max Pooling, and 5x5 Normalization. Conv1 and Conv2 are shared for all attributes. After Conv2, groupings are used to separate the layers. There are nine groups in all: \emph{Gender}, \emph{Nose}, \emph{Mouth}, \emph{Eyes}, \emph{Face}, \emph{AroundHead}, \emph{FacialHair}, \emph{Cheeks}, and \emph{Fat}. There are 6 Conv3s: one each for \emph{Gender}, \emph{Nose}, \emph{Mouth}, \emph{Eyes}, and \emph{Face}, and one for the remaining groups - Conv3Rest. Each Conv3 has 300 3x3 filters and is followed by a ReLU, 5x5 Max Pooling and 5x5 Normalization. The Conv3s are followed by fully connected layers, FC1. There are 9 FC1s - one for each group. Each FC1 is fully connected to the corresponding previous layer, with Conv3Rest connected to the FC1s for \emph{AroundHead}, \emph{FacialHair}, \emph{Cheeks}, and \emph{Fat}. Every FC1 has 512 units and is followed by a ReLU and a 50$\%$ dropout to avoid overfitting. Each FC1 is fully connected to a corresponding FC2, also with 512 units. The FC2s are followed by a ReLU and a 50$\%$ dropout. Each FC2 is then fully connected to an output node for the attributes in that group. The attributes for each group are listed below:
\begin{itemize}
  \item \textbf{Gender:} \emph{Male}
  \item \textbf{Nose:} \emph{Big Nose, Pointy Nose}
  \item \textbf{Mouth:} \emph{Big Lips, Smiling, Lipstick, Mouth Slightly Open}
  \item \textbf{Eyes:} \emph{Arched Eyebrows, Bags Under Eyes, Bushy Eyebrows, Narrow Eyes, Eyeglasses}
  \item \textbf{Face:} \emph{Attractive, Blurry, Oval Face, Pale Skin, Young, Heavy Makeup}
  \item \textbf{AroundHead:} \emph{Black Hair, Blond Hair, Brown Hair, Gray Hair, Earrings, Necklace, Necktie, Balding, Receding Hairline, Bangs, Hat, Straight Hair, Wavy Hair}
  \item \textbf{FacialHair:} \emph{5 o'clock Shadow, Mustache, No Beard, Sideburns, Goatee}
  \item \textbf{Cheeks:} \emph{High Cheekbones, Rosy Cheeks}
  \item \textbf{Fat:} \emph{Chubby, Double Chin}
\end{itemize}

The 9 groups were manually chosen according to attribute location. Some groupings were separated from others and some were absorbed into others through experimentation giving the above groupings. \emph{Male} was kept separate from all other attributes as we found, through experimentation on the CelebA dataset, that gender was improved by sharing layers with other attributes, but it ultimately decreased performance of those attributes. We found the best compromise was to include \emph{male} in the shared Conv1 and Conv2 layers and then to have separate Conv3, FC1, and FC2 layers.

We use the Caffe software for our implementation, training, and testing of MCNN and MCNN-AUX  \cite{caffe}. We use a sigmoid cross-entropy loss applied to all attribute scores to facilitate training. As preprocessing steps, the training mean is subtracted from the images and they are cropped randomly with a size of 227x227. This helps the network to be robust to shifts in the input.

If we were to use an independent CNN for each attribute following the architecture of one path in the MCNN - 3 convolutional layers and 3 fully connected layers - each CNN would have over 1.6 million parameters. So, for all 40 attributes, there would be over 64 million parameters. Using MCNN, we cut this down to less than 15 million parameters, over four times fewer.

\section{MCNN-AUX}
\label{sec:MCNN-AUX}
\begin{figure}[h]
\centering
\includegraphics[width=0.75\textwidth]{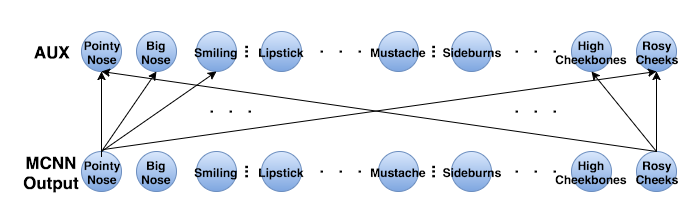}
\caption{AUX network architecture. The output of the MCNN is fully connected to the AUX layer, allowing for learning of relationships amongst attributes at the score level.}
\label{fig:aux}
\end{figure}

After training the MCNN, we add a fully connected layer, AUX, after the output of the trained MCNN. Starting with the weights from the trained MCNN, we learn the weights for the AUX portion of the network, keeping the weights from the MCNN constant. The AUX layer allows for interactions amongst attributes at the score level. The MCNN-AUX network learns the relationship amongst attribute scores in order to improve overall classification accuracy for each attribute. Figure \ref{fig:aux} shows the connection between MCNN and AUX. The AUX layer only adds 1600 parameters to the 1.6 million from MCNN.

\section{Experiments}
\label{sec:experiments}
\subsection{Data}
\begin{figure}[H]
\centering
\includegraphics[width=1.7cm]{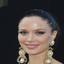}
\includegraphics[width=1.7cm]{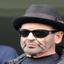}
\includegraphics[width=1.7cm]{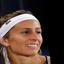}
\includegraphics[width=1.7cm]{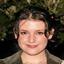}
\includegraphics[width=1.7cm]{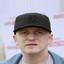}
\includegraphics[width=1.7cm]{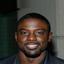}
\includegraphics[width=1.7cm]{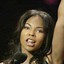}
\includegraphics[width=1.7cm]{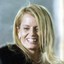}
\includegraphics[width=1.7cm]{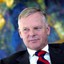}
\includegraphics[width=1.7cm]{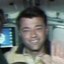}
\includegraphics[width=1.7cm]{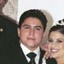}
\includegraphics[width=1.7cm]{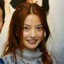}
\caption{Images from CelebA (top row) and LFWA (bottom row).}
\label{fig:data}
\end{figure}

In our experiments, we used two challenging, publicly available datasets: CelebA and LFWA. Both datasets were originally constructed for identification and verification, and recently were given binary labels for 40 different attributes \cite{Zhang12} \cite{Huang07}. Both datasets are extremely challenging, with large variations in subject pose, illumination and image quality. The CelebA dataset consists of 200,000 images: 160,000 for training and 20,000 each for validation and testing. The LFWA dataset contains 13143 images with 6263 for training and 6880 for testing. Since the CelebA dataset is so large, we did not need to augment it in any way. If we did not augment the LFWA dataset, the network would severely overfit to the training data due to the large number of parameters. We augmented the LFWA dataset by jittering the original images by increments of 10 pixels. After jittering, we had over 75,000 images for training. Figure \ref{fig:data} shows some example images from CelebA and LFWA.

\subsection{Independent CNNs}
We train independent CNNs for all the 40 attributes for both datasets in order to compare these results with those from MCNN and MCNN-AUX. We use one portion of our MCNN network for this. Each independent CNN has 3 convolutional layers, and 3 fully connected layers with the parameters specified in section \ref{sec:MCNN}. We train these networks for 22 epochs for both datasets and use a batch size of 100. The independent CNNs each take about an hour to train for the CelebA dataset and about 30 minutes for the LFWA dataset. For all 40 attributes, training independent CNNs takes over 40 hours for CelebA and over 20 hours for LFWA.

\subsection{MCNN}
To train MCNN, we use batches of size 100, and we train for 22 epochs for both datasets. Training takes about 2.5 hours for the CelebA dataset and about 1 hour for the LFWA dataset. We see a significant reduction in time from 40 hours to 2.5 hours for CelebA and 20 hours to 1 hour for LFWA using MCNN over independent CNNs.

\subsection{MCNN-AUX}
Taking the trained MCNN, we fix the weights for that portion of the MCNN-AUX network and only train the last layer, AUX. This takes about twenty minutes to train for CelebA and about 10 minutes for LFWA.

\subsection{Results}

We present results for our independent CNNs, MCNN, and MCNN-AUX. For comparison, we also provide the state-of-the-art by Liu et al., and a baseline of always choosing the most common label for each attribute.

\begin{table}
\centering
\caption{Results for CelebA. The highest accuracy for each attribute is in bold.}
\scriptsize
\begin{tabular}{| l | c | c | c | c | c |}
\hline
  \textbf{Attribute} & \textbf{Baseline} & \textbf{Liu et al.} & \textbf{Independent} & \textbf{MCNN} & \textbf{MCNN-AUX}  \\ \hline
  5 o'clock Shadow & 90.01 & 91  & 93.94 & 94.41 & \textbf{94.51} \\ \hline
  Arched Eyebrows & 71.55 & 79 & 83.16 & \textbf{83.55} &   83.42\\ \hline
  Attractive & 50.41 & 81 & 82.22 & 82.94 &  \textbf{83.06} \\ \hline
  Bags Under Eyes & 79.73 & 79 & 84.83 & 84.89& \textbf{84.92}  \\ \hline
  Bald & 97.88 & 98 & 98.85 & 98.87 &  \textbf{98.90} \\ \hline
  Bangs & 84.42 & 95 & 95.99 & 96.04 & \textbf{96.05}  \\ \hline
  Big Lips & 67.29 & 68 & 70.80 & 71.20 & \textbf{71.47}  \\ \hline
  Big Nose & 78.79 & 78 & 84.47& 84.50 &  \textbf{84.53} \\ \hline
  Black Hair & 72.83& 88 & 89.41  & \textbf{89.87} & 89.78 \\ \hline
  Blond Hair & 86.67 & 95 & 95.88& 95.97 & \textbf{96.01}  \\ \hline
  Blurry & 94.94& 84 & 96.07& 96.08& \textbf{96.17}  \\ \hline
  Brown Hair & 82.03& 80 & 88.75 & 88.99& \textbf{89.15}  \\ \hline
  Bushy Eyebrows & 87.04& 90 & \textbf{92.87} & 92.80& 92.84  \\ \hline
  Chubby & 94.69 & 91 & 95.55 & 95.66& \textbf{95.67}  \\ \hline
  Double Chin &95.42 & 92 & \textbf{96.43} & 96.41& 96.32  \\ \hline
  Earrings & 79.33& 82 & 90.35 & 90.32&  \textbf{90.43} \\ \hline
  Eyeglasses & 93.54& 99 & \textbf{99.67} & 99.63&  99.63 \\ \hline
  Goatee &95.41 & 95 & 97.13 & \textbf{97.30}& 97.24  \\ \hline
  Gray Hair & 96.81& 97 & 98.07 & \textbf{98.20}& \textbf{98.20} \\ \hline
  Hat & 95.79& 99 & 98.97& 99.04& \textbf{99.05}  \\ \hline
  Heavy Makeup & 59.50& 90 & 90.95 &91.37 & \textbf{91.55}  \\ \hline
  High Cheekbones & 51.81& \textbf{88} & 87.34 & 87.55& 87.58  \\ \hline
  Lipstick &52.18 & 93 & 93.80 & 93.95& \textbf{94.11}  \\ \hline
  Male &61.34 & 98 & 98.02& 98.16&  \textbf{98.17} \\ \hline
  Mouth Slightly Open &50.49 & 92 & \textbf{93.99} & 93.74& 93.74  \\ \hline
  Mustache &96.13 & 95 & 96.67&\textbf{96.93} & 96.88  \\ \hline
  Narrow Eyes & 85.13& 81& 87.22 & 87.16& \textbf{87.23}  \\ \hline
  Necklace & 86.20& 71 & 86.41 & \textbf{86.82}& 86.63  \\ \hline
  Necktie &92.99 & 93 & \textbf{96.71}& 96.53& 96.51  \\ \hline
  No Beard & 85.36& 95 & 95.93 & \textbf{96.11}& 96.05  \\ \hline
  Oval Face & 70.43& 66 & 74.70 & 75.81& \textbf{75.84}  \\ \hline
  Pale Skin & 95.79 & 91 & \textbf{97.07}  & 97.01& 97.05 \\ \hline
  Pointy Nose &71.42 & 72 & \textbf{77.47} & \textbf{77.47}& \textbf{77.47} \\ \hline
  Receding Hairline & 91.51& 89 & 93.41 & \textbf{93.81}& \textbf{93.81}  \\ \hline
  Rosy Cheeks & 92.82& 90 & 95.02 & 95.13& \textbf{95.16}  \\ \hline
  Sideburns &95.36 & 96 & 97.77 & 97.82& \textbf{97.85}  \\ \hline
  Smiling & 50.03& 92 & 92.65& 92.66& \textbf{92.73}  \\ \hline
  Straight Hair & 79.01& 73 &82.62  & 83.39& \textbf{83.58}  \\ \hline
  Wavy Hair &63.59 & 80 & 83.24 & \textbf{83.92}&  83.91 \\ \hline
  Young & 75.71& 87 & 87.98 & 88.30& \textbf{88.48} \\ \hline
\end{tabular}
\label{tab:resCelebA}
\end{table}

We can see from Table \ref{tab:resCelebA} that our independent CNNs outperform Liu et al. on most attributes for CelebA. The independent CNNs improve on Liu et al. by $15\%$ for \emph{necklace}, $12\%$ for \emph{blurry}, $9\%$ for \emph{straight hair}, and $8\%$ for \emph{big nose}. MCNN makes even further improvements, and finally MCNN-AUX gives the highest accuracy for most attributes. We see that the largest jump in performance is from the method of Liu et al. to the independent CNNs, with smaller improvements being made with MCNN and MCNN-AUX. From this, we see that the value in MCNN and MCNN-AUX is in the increased training speed and the decreased parameters, which reduces the chances of overfitting. We do not expect to see an increase in performance with MCNN-AUX for every attribute, as many attributes do not have strong relationships with the others. Determining which relationships to use can be done in the validation portion. We did not remove any relationships in our testing.  Unlike Liu et al., all three of our methods outperform the baseline for every attribute in CelebA.

\begin{figure}
\centering
\includegraphics[width=0.85\textwidth]{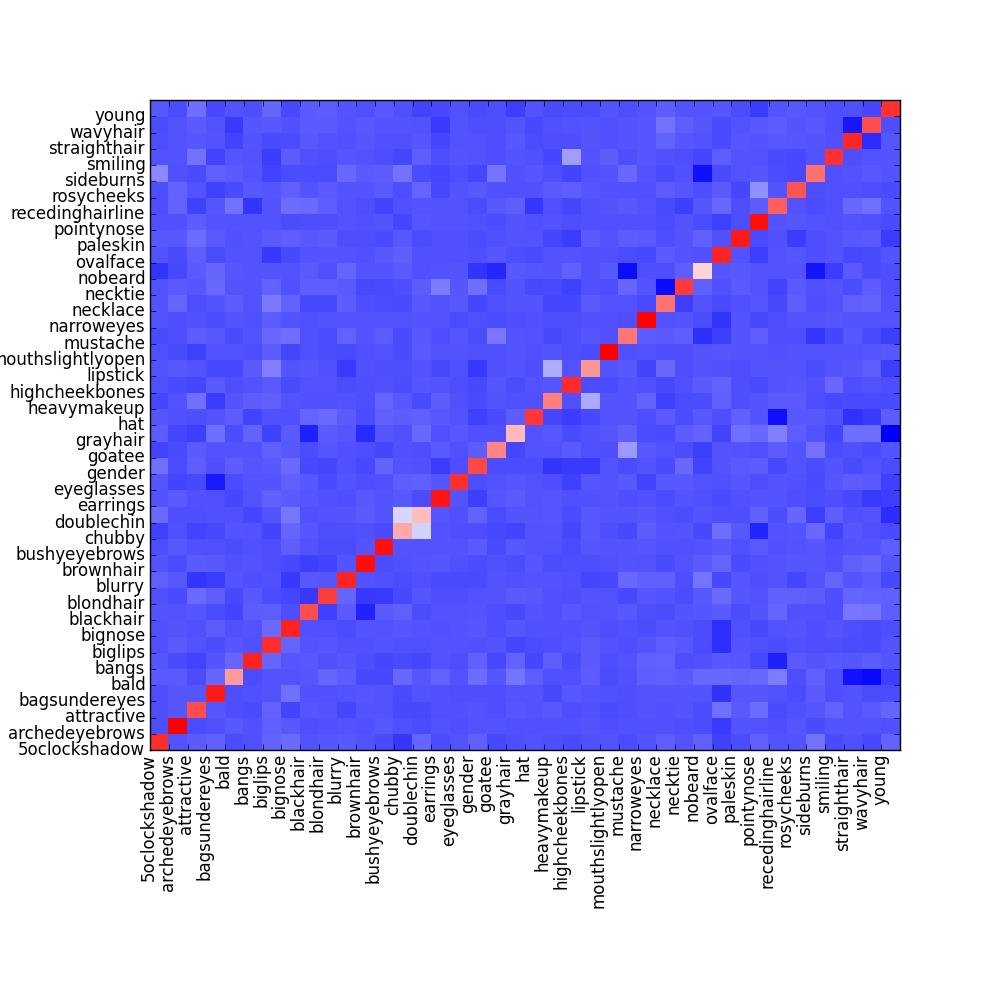}
\caption{Heatmap of AUX network weights on CelebA. Along the x-axis, we have the MCNN output units and on the y-axis, the AUX units. The red indicates a strong relationship, and the blue indicates a strong inverse relationship. Best viewed in color.}
\label{fig:heatmapCelebA}
\end{figure}

Figure \ref{fig:heatmapCelebA} shows a heatmap of the weights for the AUX layer of MCNN-AUX on the CelebA dataset. From Figure \ref{fig:heatmapCelebA} we can see that each attribute contributes the most to its final classifier score. Some interesting relationships can be seen in the heatmap. We see that \emph{bald} is strongly related to \emph{receding hairline} and has an inverse relationship with \emph{straight hair} and \emph{wavy hair} and that \emph{no beard} has an inverse relationship with \emph{5 o'clock shadow}, \emph{mustache}, and \emph{sideburns}. The strongest relationships are summarized in Table \ref{tab:celebArelations}. Most of the relationships listed in Table \ref{tab:celebArelations} make intuitive sense. Someone with \emph{heavy makeup} is likely to be wearing \emph{lipstick}; if someone is \emph{chubby}, they likely have a \emph{double chin}; and if someone has \emph{gray hair}, it is unlikely that they are \emph{young}.

\begin{table}
\centering
\caption{CelebA Attribute Relationships. N/A implies there was no strong connection for that attribute.}
\scriptsize
\begin{tabular} {| c |  C{4.5cm} | C{4.5cm} |}
\hline
\textbf{Attribute} & \textbf{Positive Influences} & \textbf{Negative Influences} \\ \hline
Bald & Receding Hairline & Straight Hair, Wavy Hair \\ \hline
Bangs & N/A & Receding Hairline \\ \hline
Black Hair & Straight Hair, Wavy Hair & Blond Hair, Brown Hair \\ \hline
Blond Hair & Attractive &  Black Hair, Brown Hair, Bushy Eyebrows \\ \hline
Chubby & Double Chin & Pointy Nose  \\ \hline
Double Chin & Chubby, Big Nose & Young \\ \hline
Eyeglasses & N/A & Bags Under Eyes \\ \hline
Male & 5 o'clock Shadow, Necktie & Earrings, Heavy Makeup, High Cheekbones, Lipstick \\ \hline
Goatee  & Mustache & 5 o'clock Shadow, No Beard  \\ \hline
Gray Hair & Receding Hairline & Black Hair, Brown Hair, Young \\ \hline
Hat & Black Hair, Blond Hair & Bald, Receding Hairline \\ \hline
Heavy Makeup & Attractive, Lipstick & Bags Under Eyes  \\ \hline
High Cheekbones & Smiling &  N/A \\ \hline
Lipstick & Heavy Makeup & Male \\ \hline
Mustache & Goatee & No Beard \\ \hline
Necklace & N/A & Necktie \\ \hline
Necktie & Male & Necklace \\ \hline
No Beard & N/A & 5 o'clock Shadow, Goatee, Male, Mustache, Sideburns \\ \hline
Receding Hairline & Bald & Bangs, Hat \\ \hline
Sideburns & 5 o'clock Shadow, Goatee & No Beard \\ \hline
Smiling & High Cheekbones & Big Lips \\ \hline
Straight Hair &N/A  & Wavy Hair \\ \hline
Wavy Hair & N/A & Straight Hair \\ \hline
Young & Attractive  & Gray Hair \\ \hline
\end{tabular}
\label{tab:celebArelations}
\end{table}

\begin{table}
\centering
\caption{Results for LFWA. The highest accuracy for each attribute is in bold.}
\scriptsize
\begin{tabular}{| l | c | c | c | c | c |}
\hline
  \textbf{Attribute} & \textbf{Baseline} & \textbf{Liu et al.} & \textbf{Independent} & \textbf{MCNN} & \textbf{MCNN-AUX}  \\ \hline
  5 o'clock Shadow & 58.64& \textbf{84} & 77.39  & 77.70 & 77.06\\ \hline
  Arched Eyebrows & 74.88& 82 & 81.4 & \textbf{82.36} & 81.78  \\ \hline
  Attractive & 62.87& \textbf{83} & 80.20 & 80.42 & 80.31   \\ \hline
  Bags Under Eyes &58.29 & 83 & 83.24 & \textbf{83.51}& 83.48   \\ \hline
  Bald &89.37 & 88 & 91.51 & \textbf{91.99}& 91.94  \\ \hline
  Bangs & 83.59& 88 & \textbf{90.47}& 89.99& 90.08  \\ \hline
  Big Lips & 62.86& 75 & 79.06& 79.21& \textbf{79.24}  \\ \hline
  Big Nose & 68.59& 81 & 84.43& 84.76&  \textbf{84.98} \\ \hline
  Black Hair & 87.63& 90 & \textbf{91.84}  & 92.35& 92.63 \\ \hline
  Blond Hair & 95.74& 97 & 97.23& \textbf{97.45}&  97.41 \\ \hline
  Blurry & 84.02& 74 & \textbf{86.71} & 85.30& 85.23  \\ \hline
  Brown Hair & 64.56& 77 & 80.84 & \textbf{80.94}& 80.85  \\ \hline
  Bushy Eyebrows & 53.70& 82 & 84.79& \textbf{85.11}& 84.97  \\ \hline
  Chubby & 63.92& 73 & 75.85& \textbf{76.90}& 76.86   \\ \hline
  Double Chin & 62.44& 78 & \textbf{82.00} & 81.17& 81.52  \\ \hline
  Earrings &86.86 & 94 & 94.73 & 94.91&\textbf{94.95}   \\ \hline
  Eyeglasses & 81.99& \textbf{95}& 92.15 & 91.22&91.30   \\ \hline
  Goatee & 74.68& 78 & \textbf{83.34} & 82.52&82.97   \\ \hline
  Gray Hair & 84.25& 84 &  88.98 & \textbf{89.04}& 88.93 \\ \hline
  Hat & 85.52& 88 & 89.79 & \textbf{90.20}& 90.07  \\ \hline
  Heavy Makeup & 89.20& 95 & 95.63 & 95.84& \textbf{95.85}  \\ \hline
  High Cheekbones &67.74 & 88 & 88.02 & 88.25&  \textbf{88.38} \\ \hline
  Lipstick &85.53 & 95 & 94.68 & 94.89& \textbf{95.04}  \\ \hline
  Male & 78.77& 94 &93.27 & 93.66& \textbf{94.02}  \\ \hline
  Mouth Slightly Open & 58.70& 82 & 82.41 & 83.47& \textbf{83.51}  \\ \hline
  Mustache & 86.62& 92 & \textbf{93.69} & 93.53& 93.43  \\ \hline
  Narrow Eyes & 65.50& 81 & 82.48 & 82.73& \textbf{82.86}  \\ \hline
  Necklace & 80.49& 88 & \textbf{89.98}& 89.66& 89.94  \\ \hline
  Necktie & 64.09& 79 & 80.34 & 80.50& \textbf{80.66}   \\ \hline
  No Beard & 70.05& 79 & 81.45 & 82.13& \textbf{82.15}  \\ \hline
  Oval Face & 51.49& 74 & 77.06. & 77.38& \textbf{77.39}  \\ \hline
  Pale Skin & 52.09& 84 & 94.31  & \textbf{93.41}& 93.32 \\ \hline
  Pointy Nose &71.10 & 80 & \textbf{84.41} & 84.18& 84.14 \\ \hline
  Receding Hairline & 59.84& 85 & 86.00 & \textbf{86.26}&  86.25 \\ \hline
  Rosy Cheeks & 79.65& 78 & \textbf{89.46} & 87.52& 87.92  \\ \hline
  Sideburns & 68.72& 77 & 81.70& 82.73&\textbf{83.13}   \\ \hline
  Smiling & 60.50& 91 & \textbf{92.22}& 91.75& 91.83  \\ \hline
  Straight Hair & 64.44& 76 & \textbf{81.54} & 78.72& 78.53  \\ \hline
  Wavy Hair & 55.49& 76 & 81.58 & \textbf{81.96}& 81.61  \\ \hline
  Young & 79.60& \textbf{86} & 85.11 & 85.37& 85.84  \\ \hline
\end{tabular}
\label{tab:resLFWA}
\end{table}

Table \ref{tab:resLFWA} shows the results for the LFWA dataset. We can see that the accuracies are lower for this dataset than for the CelebA dataset. This is likely due to overfitting because LFWA is much smaller than CelebA. The independent CNNs outperform Liu et al. on most attributes with an improvement of $11\%$ for \emph{blurry}, $11\%$ for \emph{rosy cheeks}, $10\%$ improvement for \emph{pale skin}, and $5\%$ improvements for both \emph{straight hair} and \emph{wavy hair}. MCNN improved the classification accuracy of many attributes, but we see that \emph{blurry} and \emph{eyeglasses} did not improve with MCNN. This makes sense, as both attributes are relatively unrelated to the other attributes, and therefore don't gain anything from shared information. We note that though we do not improve the results for some attributes, we perform no pre-training of the networks using a larger dataset, unlike Liu et al., which used a much larger dataset to initialize the weights of their networks. Pre-training on external data would likely improve the results, however that is not the focus of this work. 

\begin{figure}[H]
\centering
\includegraphics[width=0.85\textwidth]{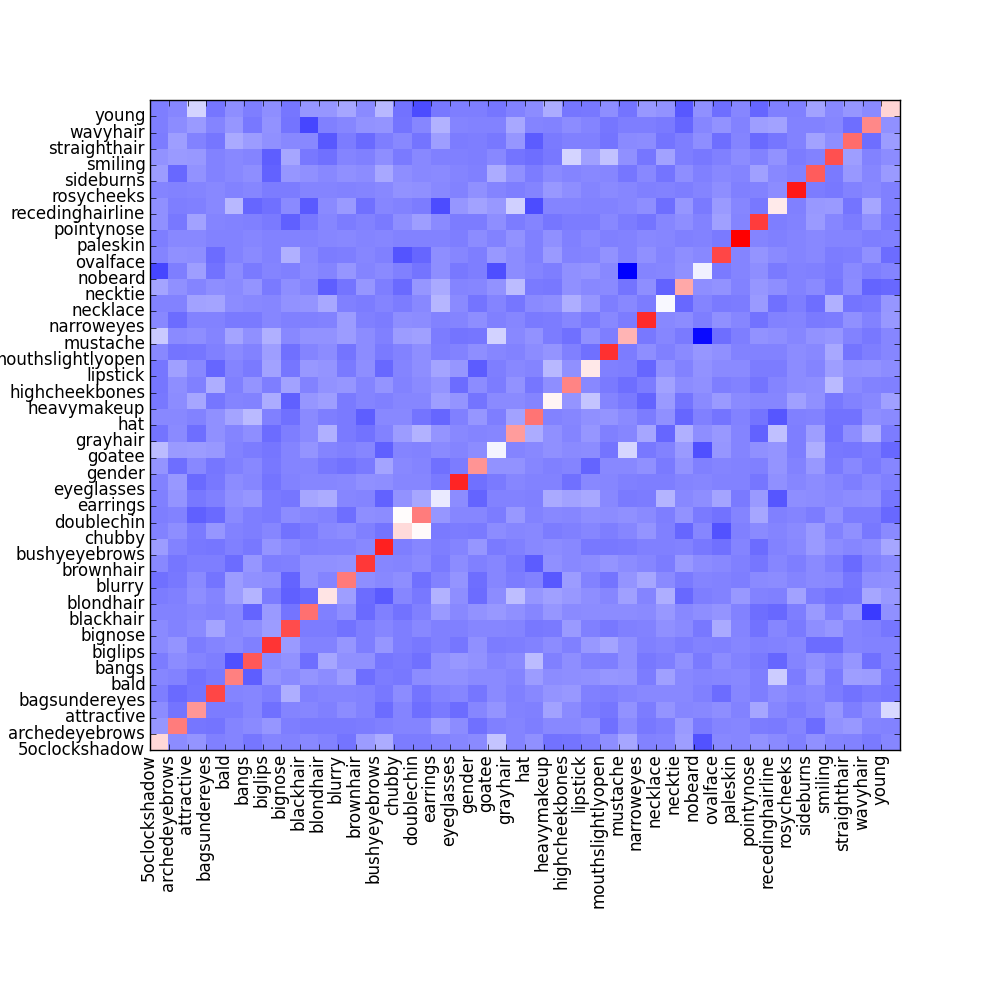}
\caption{Heatmap of AUX network weights on LFWA. Along the x-axis, we have the MCNN output units and on the y-axis, the AUX units. The red indicates a strong relationship, and the blue indicates a strong inverse relationship. Best viewed in color.}
\label{fig:heatmapLFWA}
\end{figure}

Figure \ref{fig:heatmapLFWA} shows a heatmap of the weights for the AUX layer on LFWA. There is much more white in this heatmap than in that of Figure \ref{fig:heatmapCelebA}. This makes sense, as the results for MCNN on LFWA were not as strong as on CelebA. Again, we believe that this is due to the small size of the dataset. Though jittering LFWA helps, it does not compare to having a large amount of data as in CelebA. As with CelebA, we see that each attribute contributes most to its overall classification accuracy, though not quite as strongly. We again see promising relationships, which we summarize in Table \ref{tab:lfwArelations}.

\begin{table}[H]
\centering
\caption{LFWA Attribute Relationships. N/A implies there was no strong connection for that attribute.}
\scriptsize
\begin{tabular} {| c | C{4.5cm} | C{4.5cm} |}
\hline
\textbf{Attribute} & \textbf{Positive Influences} & \textbf{Negative Influences} \\ \hline
5 o'clock Shadow & Goatee & No Beard \\ \hline
Bald & Receding Hairline & Bangs \\ \hline
Chubby & Double Chin & Oval Face \\ \hline
Double Chin & Chubby & Attractive \\ \hline
Goatee & 5 o'clock Shadow, Mustache & No Beard \\ \hline
Heavy Makeup & Lipstick & Mustache \\ \hline
High Cheekbones & Eyeglasses & Smiling \\ \hline
Lipstick & Heavy Makeup & Male \\ \hline
Mustache & Goatee & No Beard \\ \hline
No Beard & N/A & 5 o'clock Shadow, Goatee, Mustache \\ \hline
Receding Hairline & Bald, Gray Hair & Hat \\ \hline
Smiling & High Cheekbones & N/A \\ \hline
\end{tabular}
\label{tab:lfwArelations}
\end{table}


\section{Conclusion}
\label{sec:conclusion}
In this paper, we have shown that though facial attributes have been treated as independent problems in the past, there is a lot to be gained from shared information amongst attributes. Framing the attribute prediction problem as a multi-task learning problem is very natural and allows for a large reduction in training time and the number of parameters required for the classifier. In this work we showed that the MCNN-AUX reduced the number of parameters from 64 million to 16 million, and reduced the training time by 16 times. We demonstrated our independent CNN, MCNN, and MCNN-AUX classifiers on the challenging CelebA and LFWA datasets, achieving state-of-the-art performance for most attributes. The relationship amongst attributes can be exploited in many ways and we presented three ways in this paper: by sharing lower layers of MCNN, by grouping similar attributes in higher layers of MCNN, and by introducing an auxiliary layer (AUX), which explicitly learns attribute relations at the score level. Even without pre-training, we were able to outperform the method of Liu et al. for many attributes. Pre-training on external data would likely improve the results, however that is not the focus of this work. We sought to show that a multi-task framework for attribute prediction outperforms independent classifiers, and we have done that through our experimentation. Taking advantage of relationships among attributes allowed for improved attribute prediction which will lead to improved facial recognition. In future work we plan to explore how these relationships can be used to improve identification and to learn how attributes are related to identity.

\bibliographystyle{splncs}
\bibliography{egbib}
\end{document}